\documentclass{article}
\usepackage{spconf,amsmath,epsfig}
\usepackage{multirow}
\usepackage{dsfont}
\usepackage{stfloats}
\usepackage{subfig} 
\usepackage{tabu}  
\usepackage{cite}
\usepackage{booktabs}
\usepackage{algorithm}
\usepackage{algorithmic}

\title{Deep Siamese Domain Adaptation Convolutional Neural Network for Cross-domain Change Detection in Multispectral Images}
\name
 {Hongruixuan Chen$^{1}$, 
 Chen Wu$^{1}$\sthanks{Coresponding Author (chen.wu@whu.edu.cn). This work was supported in part by the National Natural Science Foundation of China under Grant 61971317, 41801285, 61822113 and 41871243.},
 Bo Du$^{2}$
 Liangpei Zhang$^{1}$
 }
 
 \address{$^1$ State Key Laboratory of Information Engineering in Surveying, Mapping, and Remote Sensing, \\
 Wuhan University, Wuhan 430079, P. R. China\\
 $^2$School of Computer, Wuhan University, Wuhan 430072, P. R. China}
 
\begin{document}
%
\maketitle
\begin{abstract}
\par Recently, deep learning has achieved promising performance in the change detection task. However, the deep models are task-specific and data set bias often exists, thus it is difficult to transfer a network trained on one multi-temporal data set (source domain) to another multi-temporal data set with very limited (even no) labeled data (target domain). In this paper, we propose a novel deep siamese domain adaptation convolutional neural network (DSDANet) architecture for cross-domain change detection. In DSDANet, a siamese convolutional neural network first extracts spatial-spectral features from multi-temporal images. Then, through multiple kernel maximum mean discrepancy (MK-MMD), the learned feature representation is embedded into a reproducing kernel Hilbert space (RKHS), in which the distribution of two domains can be explicitly matched. By optimizing the network parameters and kernel coefficients with the source labeled data and target unlabeled data, the DSDANet can learn transferrable feature representation that can bridge the discrepancy between two domains. To the best of our knowledge, it is the first time that such a domain adaptation-based deep network is proposed for change detection. The theoretical analysis and experimental results demonstrate the effectiveness and potential of the proposed method.
\end{abstract}
\section{Introduction}
\label{sec:intro}

\par Change detection (CD) is one of the most widely used interpretation techniques in the field of remote sensing, and has been intensively studied in previous years \cite{Singh1989}. Nonetheless, most traditional CD models only explore low-level features in multispectral images, which are insufficient for representing the key information of original images. Recently, deep learning (DL) has been shown to be very promising in the field of computer vision and remote sensing images interpretation. Hence, a number of CD methods based on DL models are developed. \cite{Zhu2017, Chen2019}.

\par However, the training process of these DL-based CD methods requires a lot of labeled data and there is no denying that the manual selection of labeled data is labor-consuming, especially for remote sensing images. Besides, deep networks are often task-specific, in other words, they have a relatively weak generalization. And due to several factors, including noise and distortions, sensor characteristics, imaging conditions, the data distributions of different CD data sets are often quite dissimilar. Thus, if we train a deep network on one multi-temporal data set with abundant labeled samples, it would suffer degraded performance after we transfer it to a new multi-temporal data set, which makes it unavoidable to manually label numerous samples in the new data set. Nowadays, there are massive amounts of remote sensing images are available by satellite sensors, these images can provide diverse and abundant information for covered regions. Therefore, it is incentive to develop an efficient CD model that is trained on a data set (source domain) with enough labeled data but can be easily transferred to a new data set (target domain) with very limited (even no) labeled data. This can be defined as a domain adaption problem in change detection area. 

\par Considering the above issues comprehensively, in this paper, a novel deep network architecture called DSDANet is proposed for cross-domain CD. By incorporating a domain discrepancy metric MK-MMD into the network architecture, the DSDANet can learn transferrable features, where the distribution of two domains would be similar. To the best of authors’ knowledge, it is the first time that such a deep network based on domain adaptation is designed for CD in multispectral images.

\section{Methodology}
\label{sec:methodology}
\subsection{MK-MMD}\label{sec:MKMMD}

\par Caused by plenty of factors, the probability distributions characterizing source domain $s$ and target domain $t$ are dissimilar. And due to only limited (or no) labeled data in target domain available, it is challenging to construct a model that can match these two domains and learn transferable representation. An efficient and common way is combining the CD errors with a domain discrepancy metric. 

\par A widely used metric is the maximum mean discrepancy (MMD). MMD is a nonparametric kernel-based metric that measures the distance between two distributions in a RKHS. And when the distributions of two domains tend to be the same and the RKHS is universal, MMD would approach zero.

\begin{figure*}[ht]

  \centering
  \includegraphics[scale=0.5]{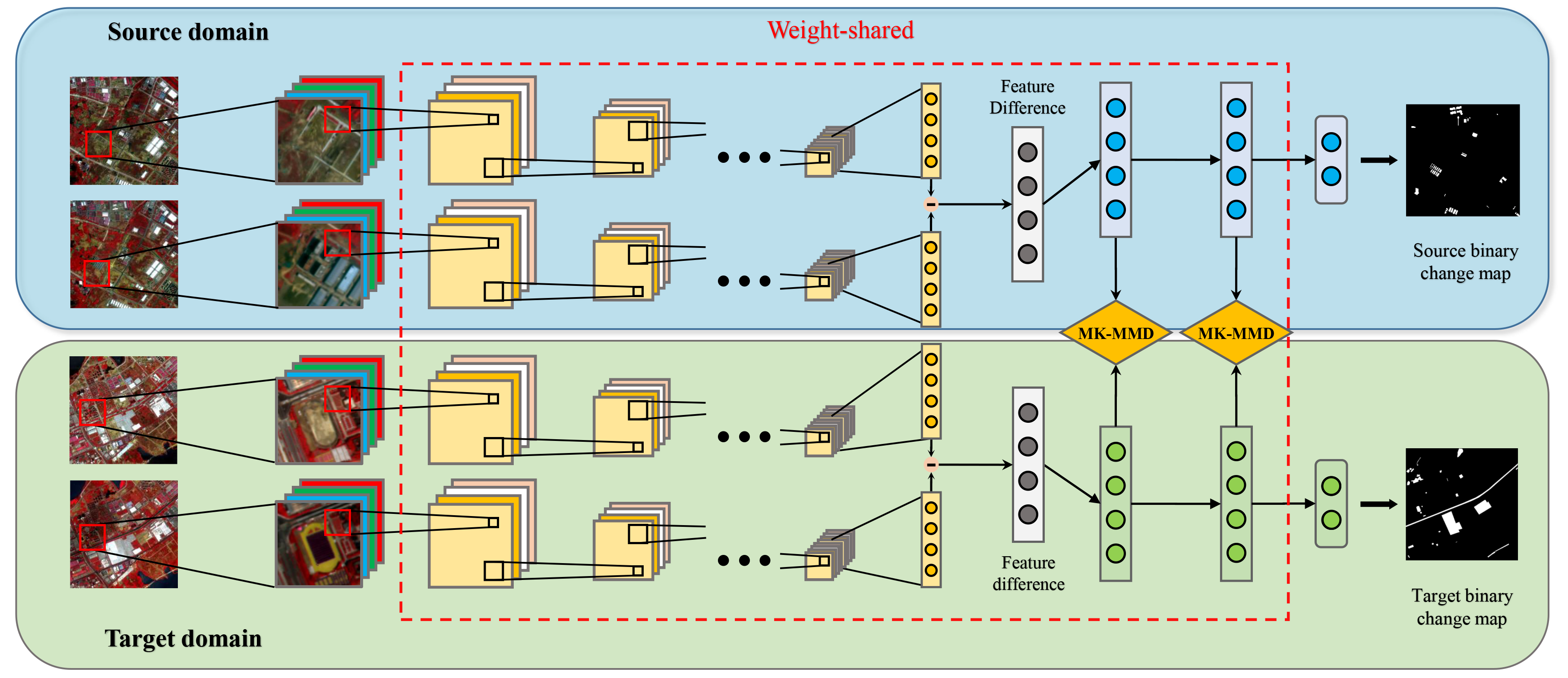}
  \caption{Overview of the CD architecture based on the proposed DSDANet.}
  \label{fig:DSDANet}
\end{figure*}

\par Nonetheless, it is difficult to find an optimal RKHS and the representation ability of single kernel is limited. And it is reasonable to assume that the optimal RKHS can be expressed as the linear combination of single kernels, thus the multi-kernel variant of MMD entitled MK-MMD \cite{Gretton2012} is introduced.

\par Considering a source data set $X_{s}$ and a target data set $X_{t}$, the formulation of MK-MMD is defined as
\begin{equation}
  d\left(X_{s},X_{t}\right)=\left\|E\left(\Phi_{k}\left(X_{s}\right)\right)-E\left(\Phi_{k}\left(X_{t}\right)\right)\right\|_{H} ,
\label{eq:2}
\end{equation}
where $\left\|\bullet\right\|_{H}$ is the RKHS norm, $\Phi_{k}\left(\bullet\right)$ is the feature map induced by multi-kernel $k$, which is defined as the linear combination of $n$ positive semi-definite kernels $\left\{k_{u}\right\}_{u=1}^{n}$
\begin{equation}
  \mathcal{K}:=\left\{k:k=\sum_{u=1}^{n}\beta_{u}k_{u},\sum_{u=1}^{n}\beta_{u}=1 ,\beta_{u} \geq 0 \right\},
\label{eq:3}
\end{equation}
where each $k_{u}$ is associated uniquely with an RKHS $H$, and we assume the kernels are bounded. Owing to leveraging diverse kernels, the representation ability of MK-MMD can get improvement. 

\par If the network can learn a domain-invariant representation that minimizes the MK-MMD between two domains, it can be easily transferred to the target domain with sparsely labeled data.

\subsection{Network Architecture}\label{sec:net_arc}
Introduced MK-MMD for domain adaptation, the structure of the proposed DSDANet is shown in Fig. \ref{fig:DSDANet}. Given a source data set $D_{s}=\left\{X_{s},Y_{s}\right\}=\left\{\left(x_{s_{i}}^{T_{1}},x_{s_{i}}^{T_{2}},y_{s_{i}}\right)\right\}_{i=1}^{n_{s}}$ with enough labeled data and a target domain $D_{t}=\left\{\left(x_{t_{i}}^{T_{1}},x_{t_{i}}^{T_{2}}\right)\right\}_{i=1}^{n_{s}}$ without labels, $x_{s_{i}}^{T_{n}}\in R^{k_{1}\times k_{2}\times c}$ is an image patch centered $i$-th pixel and $y_{s_{i}}$ is the corresponding label of $i$-th pixel. For each image patch-pair in both domains, the spatial-spectral features $f_{i}^{T_{1}}$ and $f_{i}^{T_{2}}$ are extracted by cascade convolutional layers and max-pooling layers.

\par After that, the absolute value of multi-temporal spatial-spectral features’ difference is calculated. Since the two branches of DSDANet are weight-shared, the change information could be highlighted through this operation. 

\par As we all konw, deep features learned by CNN transition from general to specific by the network going deeper. Especially for the last few fully connected (FC) layers, there exists an insurmountable transferability gap between features learned from different domains. If we train a network in the source domain, it cannot be transferred to the target domain via fine-tuning with sparse target labeled data. Therefore, the MK-MMD is adopted to make the network learn domain-invariant features from two domains. An intuitive idea is combining MK-MMD with the penultimate FC layer, which can directly make the classifier adaptive to two domains. But considering a single layer may not cope with domain distribution bias, thus the MK-MMD is embedded into the two FC layers in front of the classifier. Since we aim to construct a network that is trained on the source CD data set but also perform well on the target task, thus the loss function of DSDANet is
\begin{equation}
  \mathcal{L}=\mathcal{L}_{C}\left(X_{s},X_{t}\right)+\lambda \sum_{l=l_{a}}^{l_{a}+1}d_{k}^{2}\left(D_{s}^{l},D_{t}^{l}\right),
  \label{eq:5}    
\end{equation}
where $\mathcal{L}_{C}\left(X_{s},X_{t}\right)$ is CD loss on the source labeled data, $l_{a}$ is layer index, $d_{k}\left(D_{s}^{l},D_{t}^{l}\right)$ means the MK-MMD between the two domain on the features in the $l$-th layer and $\lambda \ge 0$ denotes a domain adaptation penalty parameter.

\subsection{Optimization}\label{sec:opt}
\par In the training procedure, two types of parameters require to learn, one is the network parameters $\Theta$ and another is the kernel coefficient $\beta$. However, the cost of MK-MMD computation by kernel trick is $O\left(n^{2}\right)$, it is unacceptable for deep networks in large-scale data sets and makes the training procedure more difficult. Therefore, the unbiased estimate of MK-MMD \cite{Gretton2012} is utilized to decrease the computation cost from $O\left(n^{2}\right)$ to $O\left(n\right)$, which can be formulated as
\begin{equation}
  \left\{
    \begin{aligned}
      &d_{k}^{2}\left(D_{s}^{l},D_{t}^{l}\right)=\frac{2}{n_{s}}\sum_{i=1}^{n_{s}}g_{k}\left(z_{i}^{l}\right) \\
      &g_{k}\left(z_{i}\right)=k\left(h_{2i-1}^{sl},h_{2i}^{sl}\right)+k\left(h_{2i-1}^{tl},h_{2i}^{tl}\right)\\
      &\qquad\qquad -k\left(h_{2i-1}^{sl},h_{2i}^{tl}\right)-k\left(h_{2i}^{sl},h_{2i-1}^{tl}\right)
    \end{aligned}
  \right.  
  \label{eq:6}    
\end{equation}
where $z_{i}^{l}=\left(h_{2i-1}^{sl},h_{2i}^{sl},h_{2i-1}^{tl},h_{2i}^{tl}\right)$ is a quad-tuple evaluated by multi-kernel $k$ and $h^{l}$ is learned features in $l$-th layer. 

\par As for the kernel parameters $\beta$, the optimal coefficient for each $d_{k}^{2}\left(D_{s}^{l},D_{t}^{l}\right)$ can be sought by jointly maximizing $d_{k}^{2}\left(D_{s}^{l},D_{t}^{l}\right)$ itself and minimizing the variance, which results in the optimization
\begin{equation}
  \max\limits_{k\in\mathcal{K}}d_{k}^{2}\left(D_{s}^{l},D_{t}^{l}\right)/\sigma_{k}^{2},
  \label{eq:7}    
\end{equation}
where $\sigma_{k}^{2}$ is estimation variance. Eventually, this optimization finally can be resolved as a quadratic program (QP) \cite{Gretton2012}.

\par By alternatively adopting stochastic gradient descent (SGD) to update $\Theta$ and solving QP to optimize $\beta$, the DSDANet can gradually learn transferrable representation from source labeled data and target unlabeled data. By minimizing Eq. \ref{eq:5}, the marginal distributions $P\left(X_{s}\right)$ and $P\left(X_{t}\right)$ of two domains become very similar, yet the conditional distributions $P\left(Y_{s}|X_{s}\right)$ and $P\left(Y_{t}|X_{t}\right)$ of two domains may still be slightly different. Thus, a very small part of target labeled data is selected to fine-tune the classifier of DSDANet. Compared with the enough labeled data in the source domain, the labeled data provided by the target domain is very limited, so this procedure can be treated as a semi-supervised learning fashion.

\begin{figure}[t]
  \centering
  \subfloat[]{
    \includegraphics[width=0.9in]{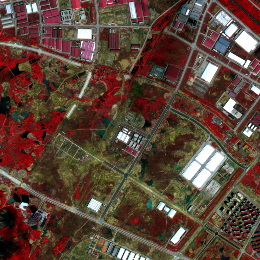}
  \label{WH_1}}
  \hfil
  \subfloat[]{
    \includegraphics[width=0.9in]{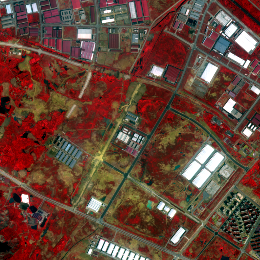}
  \label{WH_2}}
  \hfil
  \subfloat[]{
    \includegraphics[width=0.9in]{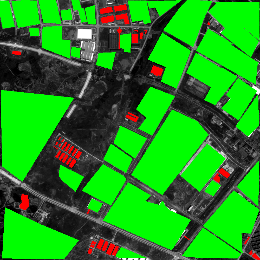}
  \label{WH_GT}}
  \caption{The WH data set adopted as source domain. In the ground truth, red indicates change and green means non-change.}
  \label{source}
\end{figure}

\begin{figure}[t]
  \centering
  \subfloat[]{
    \includegraphics[width=0.9in]{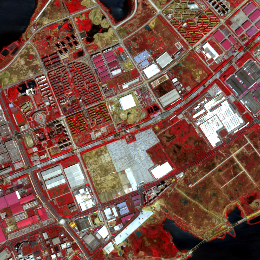}
  \label{HY_1}}
  \hfil
  \subfloat[]{
    \includegraphics[width=0.9in]{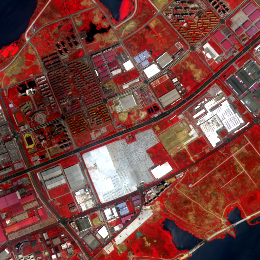}
  \label{HY_2}}
  \hfil
  \subfloat[]{
    \includegraphics[width=0.9in]{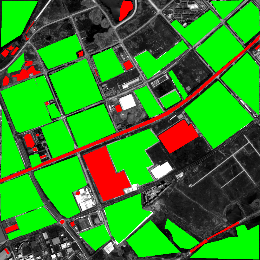}
  \label{HY_GT}}
  
  \subfloat[]{
    \includegraphics[width=0.9in]{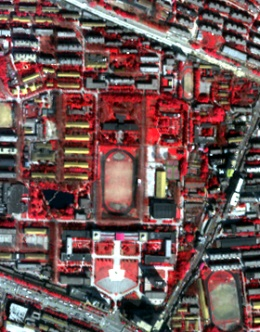}
  \label{QU_1}}
  \hfil
  \subfloat[]{
    \includegraphics[width=0.9in]{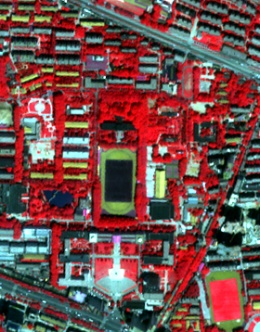}
  \label{QU_2}}
  \hfil
  \subfloat[]{
    \includegraphics[width=0.9in]{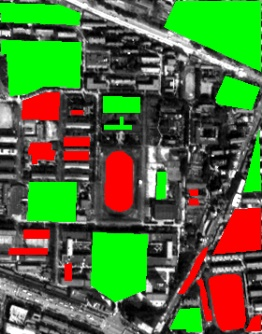}
  \label{QU_GT}}
  \caption{Two data sets adopted as target domains. (a)-(c) HY data set. (d)-(f) QU data set. In the ground truth, red indicates change and green means non-change.}
  \label{target}
\end{figure}

\section{Experiments}
\label{sec:experiments}
\subsection{General Information}\label{sec:general}

\par The data set used as the source domain is WH data set captured by GF-2, as shown in Fig. \ref{source}. The size of the two images is 1000 $\times$ 1000 pixels with four spectral bands and they have a spatial resolution of 4m. 

\par The data sets adopted as the target domains are HY data set and QU data set, as shown in Fig. \ref{target}. The HY data set was also captured by GF-2 with a size of 1000 $\times$ 1000 pixels. The second target data set was acquired by QuickBird with four spectral bands and a spatial resolution of 2.4m denoted as QU. Both images in this data set are 358 $\times$ 280 pixels. Since the WH and QU were acquired by different sensors leading to diverse spatial resolutions and statistical characteristics, the data distributions of these two data sets are significantly different. 

\par In the training procedure, we randomly select 10$\%$ samples (the particular number is 50416) from the source domain as labeled training samples. And we train the DSDANet with labeled source training samples and all target samples without labels. After training, we only select 200 labeled samples from each target domain for fine-tuning the classifier. Compared with the labeled source data, the labeled data provided by the target domain is sparse.

\par To evaluate the proposed method, we compare it with CVA \cite{Sharma2007} and SVM. To further evaluate the effectiveness of MK-MMD, we compare the DSDANet to its variants that don’t perform domain adaptation, including directly inferring target data without fine-tuning (DSCNet-v1), directly training in the target labeled data instead of training in the source domain (DSCNet-v2) and fine-tuning with target labeled data but not equipped with MK-MMD (DSCNet-v3).

\begin{figure}[t]
  \centering
  \subfloat[]{
    \includegraphics[width=0.49in]{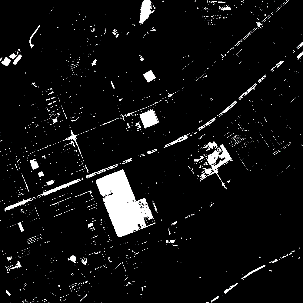}
  \label{HY_MAD}}
  \hfil
  \subfloat[]{
    \includegraphics[width=0.49in]{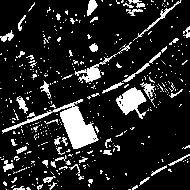}
  \label{HY_IRMAD}}
  \hfil
  \subfloat[]{
    \includegraphics[width=0.49in]{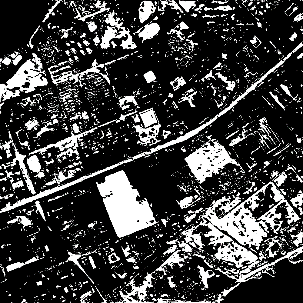}
  \label{HY_USFA}}
  \hfil
  \subfloat[]{
    \includegraphics[width=0.49in]{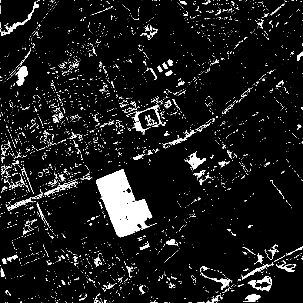}
  \label{HY_ISFA}}
  \hfil
  \subfloat[]{
    \includegraphics[width=0.49in]{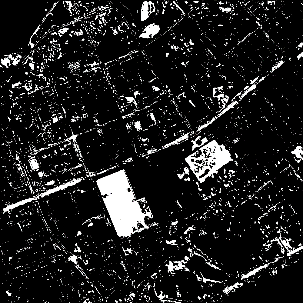}
  \label{HY_CVA}}
  \hfil
  \subfloat[]{
    \includegraphics[width=0.49in]{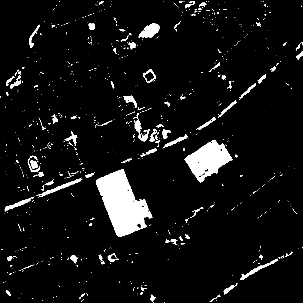}
  \label{HY_OBCD}}
  \caption{Binary change maps obtained by the proposed method and comparison methods on the WH. (a) CVA. (b) SVM. (c)-(e) Variants of DSDANet. (f) DSDANet.}
  \label{HY_result}
\end{figure}

\begin{figure}[t]
  \centering
  \subfloat[]{
    \includegraphics[width=0.49in]{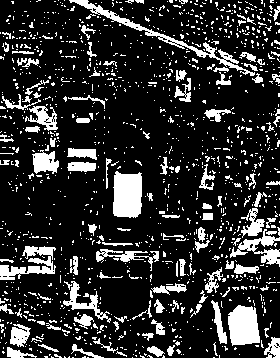}
  \label{QU_MAD}}
  \hfil
  \subfloat[]{
    \includegraphics[width=0.49in]{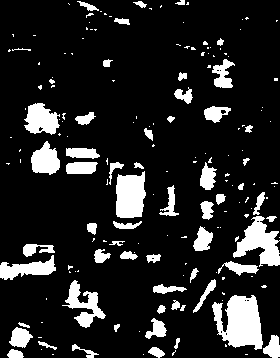}
  \label{QU_IRMAD}}
  \hfil
  \subfloat[]{
    \includegraphics[width=0.49in]{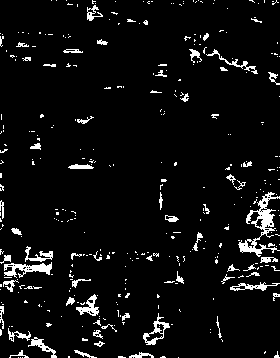}
  \label{QU_USFA}}
  \hfil
  \subfloat[]{
    \includegraphics[width=0.49in]{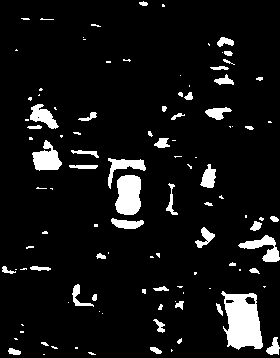}
  \label{QU_ISFA}}
  \hfil
  \subfloat[]{
    \includegraphics[width=0.49in]{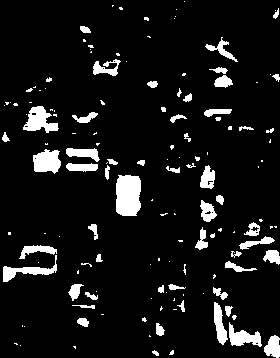}
  \label{QU_CVA}}
  \hfil
  \subfloat[]{
    \includegraphics[width=0.49in]{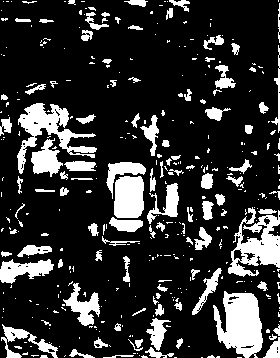}
  \label{QU_OBCD}}
  \caption{Binary change maps obtained by the proposed method and comparison methods on the QU. (a) CVA. (b) SVM. (c)-(e) Variants of DSDANet. (f) DSDANet.}
  \label{QU_result}
\end{figure}

\subsection{Experimental Results}\label{sec:experi_result}

\par The binary change maps obtained by different methods on the HY data set are shown in Fig. \ref{HY_result}. It can be observed that the proposed model generates the best CD result with more complete changed regions and less noise. For the QU data set, even though the distributions of the two domain are significantly different due to the diverse characteristics of the two sensors, the DSDANet still can generate an accurate binary change map. It implies that through embedding data distributions into the optimal RKHS and minimize the distance between them, the network is capable of learning domain-invariant representation from source labeled data and unlabeled target data and can be easily transferred from one CD data set to another. 

\par The quantitative results are listed in Table \ref{table_1}. Due to only providing very limited target labeled data that cannot contain all the kinds of changed and unchanged land-cover types, fine-tuning without domain adaptation also performs not well. By contrast, the DSDANet achieves the best OA and KC on the two target data set.

\begin{table}[t]\footnotesize
  \renewcommand{\arraystretch}{1.2}
  \caption{Accuracy assessment on binary change maps obtained by different methods on the two target data set}
  \label{table_1}
  \centering
  \begin{tabular}{c c c c c}
    \hline
    \multirow{2}{*}{\textbf{Method}} & \multicolumn{2}{c}{\textbf{HY}} & \multicolumn{2}{c}{\textbf{QU}}  \\
    \cline{2-5}
     &  \textbf{OA} &  \textbf{KC} &  \textbf{OA} & \textbf{KC} \\
    \hline\hline
    CVA	& 0.9445	& 	0.7171	& 0.8079 & 0.5352	\\ 	
    SVM	& 0.8467	& 	0.4565	& 0.8381 & 0.6285	\\ 	
    DSCNet-v1	& 0.8751	& 	0.4310	& 0.7060 & 0.1147	\\ 	
    DSCNet-v2	& 0.8759	& 	0.5610	& 0.8286 & 0.5404	\\ 	
    DSCNet-v3	& 0.9279	& 	0.6650	& 0.8297 & 0.5391	\\
    DSDANet	& \textbf{0.9618}	& 	\textbf{0.8021}	& \textbf{0.9016} & \textbf{0.7670}	\\ 	
    \hline
  \end{tabular}
\end{table}

\section{Conclusion}
\label{sec:conclusion}

\par In this paper, a novel network architecture entitled DSDANet is proposed for cross-domain CD in multispectral images. Through restricting the domain discrepancy with MK-MMD and optimizing the network parameters and kernel coefficient, the DSDANet can learn transferrable representation from source labeled data and target unlabeled data, which can efficiently bridge the discrepancy between two domains. The experimental results in two target data sets demonstrate the effectiveness of the proposed DSDANet in cross-domain CD. Even though the data distributions of the two domains are significantly different, the DSDANet only needs sparse labeled data of the target domain to fine-tune the classifier, which makes it superior in actual production environments.


\bibliographystyle{IEEEbib}
\bibliography{strings,refs}

\end{document}